\documentclass[letterpaper,10pt,conference]{ieeeconf}
\IEEEoverridecommandlockouts
\usepackage[T1]{fontenc}
\usepackage{graphicx,amsmath,amssymb,booktabs,array,tabularx,cite,url}
\usepackage[hidelinks]{hyperref}
\hypersetup{
  pdftitle={GeoBridge-VLA: Geometry-Aware Residual Adaptation for Vision-Language-Action Models},
  pdfauthor={Hyun Song, Kangmin Kim, Loren Jinsoo Um, Minhui Han, Jaehyeok Park, Taewan Cho, Andrew Jaeyong Choi},
  pdfsubject={Geometry-aware adaptation for vision-language-action models},
  pdfkeywords={GeoBridge-VLA, vision-language-action models, robot learning, geometry},
  pdflang={en-US}
}
\usepackage{placeins}
\usepackage{capt-of}
\newcommand{\method}{GeoBridge-VLA}

\title{\LARGE\bf GeoBridge-VLA: Geometry-Aware Residual Adaptation for Vision-Language-Action Models}
\author{Hyun Song, Kangmin Kim, Loren Jinsoo Um, Minhui Han,\\
Jaehyeok Park, Taewan Cho, and Andrew Jaeyong Choi$^{*}$\\[0.8ex]
{\normalsize School of Computing, iRASC Lab., Gachon University}\\[0.3ex]
{\small\texttt{\{songhyun, kgmin2413, jinsoo7942, kellen0323\}@gachon.ac.kr}}\\[0.1ex]
{\small\texttt{\{joonwall09, taewan2002, andrewjchoi\}@gachon.ac.kr}}%
\thanks{\protect\raggedright $^{*}$Corresponding author: Andrew Jaeyong Choi.}%
}
\IEEEaftertitletext{%
\begin{minipage}{\textwidth}
\centering
\includegraphics[width=\textwidth]{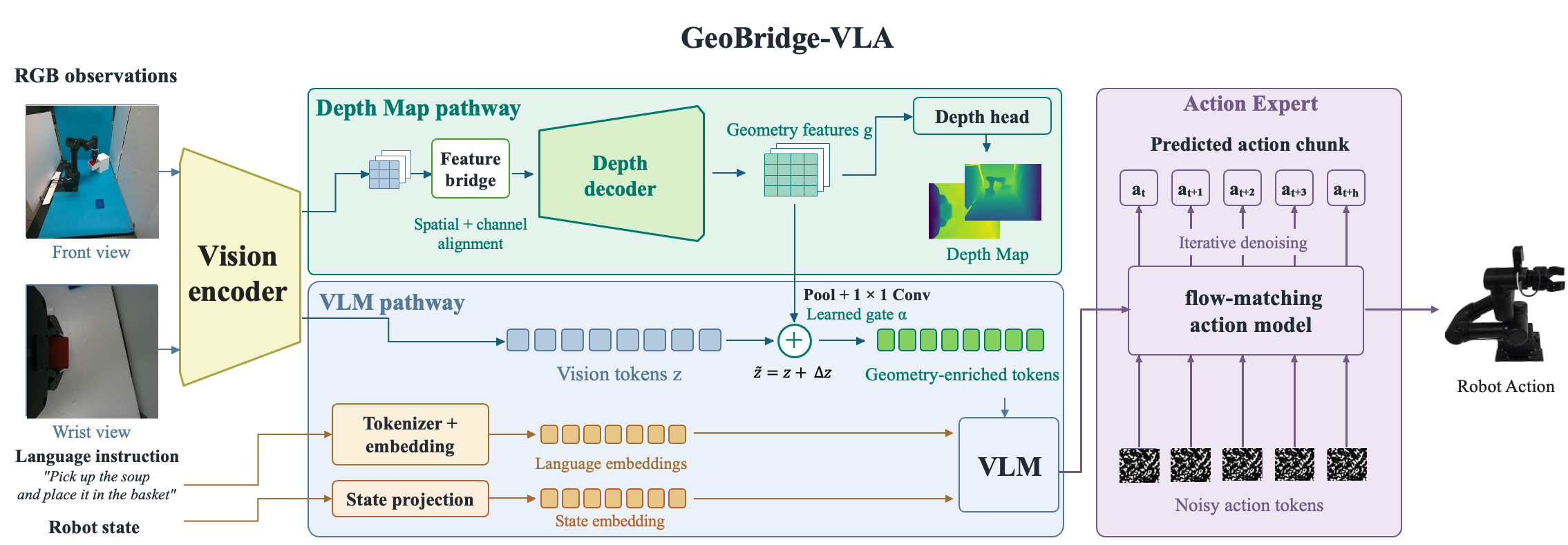}
\captionof{figure}{\textbf{Overview of GeoBridge-VLA.} Each RGB view passes through the shared native vision encoder. Selected intermediate features enter the feature bridge and K4-based geometry decoder, producing a depth map and geometry features $g$. Pooling, a $1\times1$ readout, and a learned gate transform $g$ into a residual $\Delta z$ on the native visual tokens. The existing conditioning pathway combines the adapted visual representation with language and robot state for action generation. Stage I trains the bridge and decoder; Stage II freezes them and trains the residual interface and action components.}
\label{fig:method}
\vspace{1em}
\end{minipage}%
}
\begin{document}
\maketitle
\thispagestyle{empty}
\pagestyle{empty}
\begin{figure*}[t]
\centering
\includegraphics[width=0.88\textwidth]{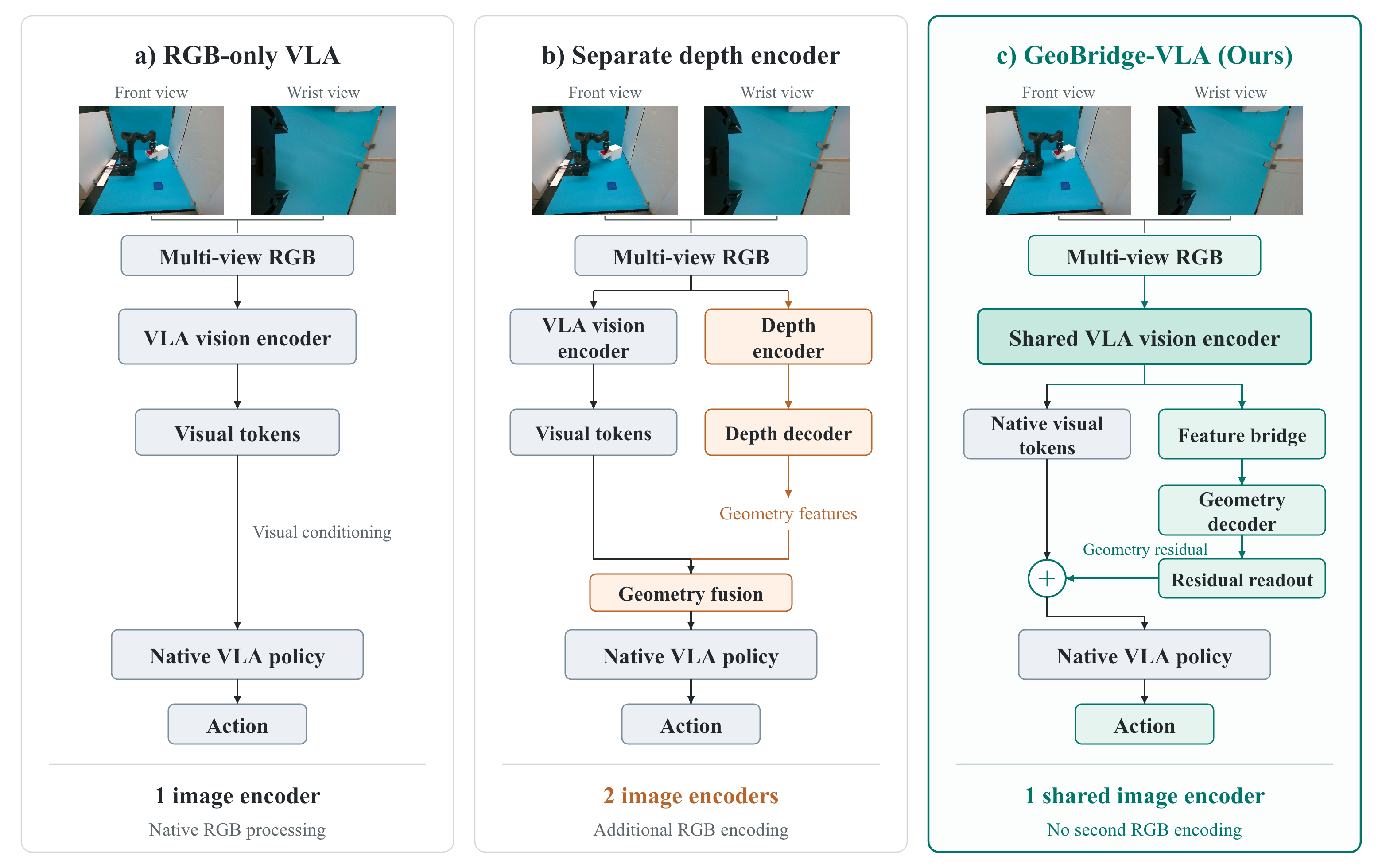}
\caption{\textbf{Geometry integration in VLA policies.} (a) A native RGB-only VLA uses its visual tokens for action conditioning. (b) A separate-depth design adds a second RGB encoder and a fusion module. (c) \method{} shares the native visual encoder, decodes geometry through a feature bridge and K4-based geometry decoder, and adds a learned residual to the existing visual tokens. Language and robot-state inputs are omitted here for clarity.}
\label{fig:concept}
\end{figure*}
\begin{abstract}

Vision-language-action (VLA) models encode semantic information from
vision-language pretraining, but manipulation also requires precise spatial
reasoning. We present GeoBridge-VLA, a two-stage method for learning
geometric features from a pretrained VLA's frozen visual encoder and using
them for action prediction. Stage I trains a feature bridge and geometry
decoder with depth supervision. Stage II freezes these modules and trains
a gated residual interface together with the action-side projections and
action expert. The residual augments the existing visual tokens without
adding a second image encoder or increasing the token count. Deployment
requires RGB, robot state, and language, but no depth observations.
Under matched evaluation conditions, GeoBridge-VLA achieves 70.9\% success
on LIBERO, compared with 60.0\% for SmolVLA. Disabling the residual in the
same trained checkpoint reduces success from 70.90\% to 69.85\%, with
mixed effects across suites. On a physical ROBOTIS OMY robot, GeoBridge-VLA
succeeds in 148 of 200 trials (74.0\%) across four tasks, compared with
108 of 200 (54.0\%) for SmolVLA.

\end{abstract}

\section{INTRODUCTION}
\label{sec:introduction}

Vision-language-action (VLA) models extend pretrained vision-language
representations to language-conditioned robotic manipulation
\cite{rt2,openvla,pi0,smolvla}. Their pretrained representations provide
strong semantic priors for recognizing objects, interpreting instructions,
and generalizing across diverse scenes. Robotic manipulation, however, also
depends on geometric properties such as depth, relative position, object
extent, and local spatial relationships. Representations optimized primarily
for semantic alignment do not necessarily preserve this information with the
precision required for physical interaction.

Recent work has addressed this limitation through explicit depth or 3D
observations, external spatial representations, calibrated camera geometry,
and auxiliary geometric supervision. SpatialVLA introduces 3D-aware spatial
representations~\cite{spatialvla}, DepthVLA incorporates a pretrained depth
module~\cite{depthvla}, FALCON injects features from a spatial foundation
model~\cite{falcon}, G$^3$VLA introduces calibrated camera geometry into the
visual-token stream~\cite{g3vla}, and GIFT shapes intermediate policy
representations through control-oriented structural supervision
~\cite{gift}. We investigate whether the frozen visual hierarchy of a VLA
can provide depth-supervised features for action prediction without a
separate image encoder, additional geometry tokens, or depth observations
at deployment.

We refer to this adaptation as \emph{geometry retrofit}. GeoBridge-VLA
first learns a geometric readout from frozen intermediate visual features,
then transfers the resulting features to the action policy. Here, geometry
refers to the representation learned through depth supervision. In Stage I,
a feature bridge maps selected encoder features to a geometry decoder.
We train the bridge and decoder using simulator ground-truth depth for
LIBERO and teacher-generated pseudo-depth for OMY. In Stage II, we freeze
both modules and train a gated residual interface that adds decoder
features to the native visual tokens during action learning. The two
paths share the VLA's visual encoder, preserving the token count and
existing conditioning pathway.

We evaluate GeoBridge-VLA across all four LIBERO manipulation suites and on a
physical ROBOTIS OMY platform. The complete adaptation pipeline improves
manipulation success over the SmolVLA baseline in simulation and physical
robot evaluation, while a same-checkpoint residual intervention tests the
trained policy's sensitivity to the learned geometry-to-action pathway.
Detailed results are reported in Sec.~\ref{sec:results}.

The main contributions of this work are:
\begin{itemize}
    \item We formulate geometry integration in pretrained VLAs as a
    \emph{geometry retrofit} problem and introduce a shared-hierarchy pathway
    that learns a depth-supervised geometric readout directly from frozen
    intermediate VLA features without adding a separate image encoder to the
    deployed policy.

    \item We propose a two-stage geometry acquisition and geometry-to-action
    transfer strategy in which depth supervision first adapts the feature
    bridge and geometry decoder, after which they are frozen and a gated
    residual interface transfers their representation into the native
    visual-token space for action learning.

    \item We evaluate the resulting policy across all four LIBERO suites and
    on a physical ROBOTIS OMY robot under depth-free deployment, and further
    analyze shared-encoder depth prediction and same-checkpoint residual
    intervention to examine geometry acquisition and the trained policy's
    sensitivity to geometry-to-action transfer.
\end{itemize}

\section{Related Work}
\label{sec:related-work}

\subsection{Monocular Depth Estimation}

Transformer-based monocular depth estimation has progressed from DPT and
ZoeDepth to large-scale foundation models such as Depth Anything V2 and
Depth Anything 3~\cite{dpt,zoedepth,depthanythingv2,da3}. CLIP-based
approaches further demonstrate that depth-related information can be
recovered from frozen vision--language representations: DepthCLIP exploits
patch--text correlations, while SPACE-CLIP directly decodes semantic and
structural features from a frozen CLIP vision encoder
~\cite{depthclip,spaceclip}. These works primarily target depth prediction;
GeoBridge-VLA instead uses depth supervision to adapt geometric information
for closed-loop action generation.

\subsection{Vision-Language-Action Models}

VLA models extend pretrained vision--language representations to robotic
control. RT-2 and OpenVLA demonstrate large-scale vision--language-to-action
transfer~\cite{rt2,openvla}, while $\pi_0$ uses flow matching for continuous
action generation~\cite{pi0}. Recent systems also emphasize efficient policy
design and deployment, including OpenVLA-OFT, TinyVLA, and SmolVLA
~\cite{openvlaoft,tinyvla,smolvla}. We use SmolVLA as the policy backbone
throughout our action-policy experiments.

\subsection{Geometry-Aware Vision-Language-Action Models}

Recent VLA systems incorporate geometric or spatial structure through several
mechanisms. Explicit 3D approaches use depth, point clouds, or dedicated
spatial representations: 3D-VLA couples 3D perception and action generation
~\cite{3dvla}, FP3 and PointVLA operate on point-cloud features
~\cite{fp3,pointvla}, and 3D-CAVLA combines depth perception with
task-oriented spatial reasoning~\cite{3dcavla}. Other methods improve spatial
reasoning without requiring metric depth at deployment. IVRA reuses affinity
information from the native visual encoder at inference~\cite{ivra}, while
RetoVLA repurposes Register Tokens for spatial reasoning~\cite{retovla}.
FALCON injects features from an embodied spatial model~\cite{falcon}, and
GIFT imposes geometry-aware structural supervision on intermediate policy
representations~\cite{gift}.

Other approaches incorporate geometric structure into VLA policies through
different mechanisms. SpatialVLA predicts depth and back-projects image
features into egocentric 3D structure~\cite{spatialvla}. G$^3$VLA injects
calibrated camera geometry through geometric encodings and cross-view fusion
~\cite{g3vla}, while GP3 and AugVLA-3D introduce dedicated spatial encoders
or VGGT-derived 3D features before action prediction~\cite{gp3,augvla3d}.
QDepth-VLA and DepthVLA incorporate dedicated depth experts or pretrained
depth transformers~\cite{qdepthvla,depthvla}, whereas Evo-Depth extracts
implicit depth features from RGB and progressively aligns depth-aware
modulation with action learning~\cite{evodepth}. Concurrent GaussVLA lifts
semantic and depth features into compact 3D Gaussian tokens for geometric
reasoning~\cite{gaussvla}.

GeoBridge-VLA instead learns a depth-supervised geometric readout directly
from intermediate features of the frozen native VLA visual encoder and
transfers the resulting representation through a gated residual interface.
This separates geometry acquisition from geometry-to-action transfer while
requiring neither a second image encoder nor additional visual tokens at
deployment.

\section{Method}
\label{sec:method}

GeoBridge-VLA incorporates depth-derived geometric features into a VLA's
existing visual conditioning pathway. Let $F^\ell$ denote selected
intermediate visual features and $z$ the native final visual tokens. As
illustrated in Fig.~\ref{fig:method}, we first adapt a feature bridge and 
K4-based geometry decoder to the frozen visual encoder, then learn a 
geometric residual interface for action prediction.

\subsection{Stage I: Depth Model Adaptation}

\textbf{Feature bridge.}
We adopt a geometry decoder adapted from SPACE-CLIP~\cite{spaceclip}
with fixed local token aggregation, termed K4. At each aggregation
stage, the decoder combines features sampled at four fixed neighboring
positions using content-dependent weights, preserving local spatial
structure. This design improves standalone depth prediction over
the original decoder under matched backbone and data settings.
The formulation and standalone comparison are provided in the
\hyperref[app:decoder]{appendix}.
We connect this decoder to the native VLA features through a learnable
bridge. Each selected feature is spatially
resampled by $\mathcal{S}$ and transformed by a layer-specific linear
projection $B_\ell$, applied independently at each location:
\begin{equation}
\begin{aligned}
H^\ell &= B_\ell\bigl(\mathcal{S}(F^\ell)\bigr),\\
(\widehat d,g) &= D_{\mathrm{geo}}(H^{\ell_1},\ldots,H^{\ell_L}).
\end{aligned}
\label{eq:bridge}
\end{equation}
Here $D_{\mathrm{geo}}$ denotes the K4-based geometry decoder, $\widehat d$
is its depth prediction, and $g$ is the feature map used for action
conditioning. The bridge aligns the spatial and channel dimensions of the
selected features. Although this interface can accommodate compatible
visual backbones, our policy experiments use SmolVLA only. The action
interface receives $g$ directly; it does not re-encode the predicted depth
map.

\textbf{Depth adaptation.}
We optimize the bridge and geometry decoder while keeping the native visual
encoder frozen. The objective combines scale-invariant log-depth
supervision~\cite{eigen2014depth} and structural similarity
~\cite{wang2004ssim} with auxiliary depth supervision:
\begin{equation}
\begin{aligned}
\mathcal{L}_{\mathrm{I}}
={}&
\lambda_{\mathrm{SILog}}
\mathcal{L}_{\mathrm{SILog}}(\widehat d,d)
+\lambda_{\mathrm{SSIM}}
\mathcal{L}_{\mathrm{SSIM}}(\widehat d,d)\\
&+\sum_{s=1}^{S}\lambda_s
\mathcal{L}_{\mathrm{SILog}}(\widehat d_s,d).
\end{aligned}
\label{eq:depth}
\end{equation}
Here $\widehat d_s$ denotes an auxiliary prediction at scale $s$, and
$\lambda_{\mathrm{SILog}}$, $\lambda_{\mathrm{SSIM}}$, and $\lambda_s$ are
fixed loss-balancing coefficients. The target $d$ is valid simulator
ground-truth depth for LIBERO and teacher-generated pseudo-depth for OMY.
For LIBERO, adaptation uses no teacher model or reference encoder. For OMY,
a frozen Depth Anything 3~\cite{da3} teacher provides the Stage-I depth targets, as
described in Sec.~\ref{sec:omy-setup}. The teacher is used only to provide
training supervision and is absent from the deployed policy.

\begin{figure*}[t]
\centering
\includegraphics[width=\textwidth]{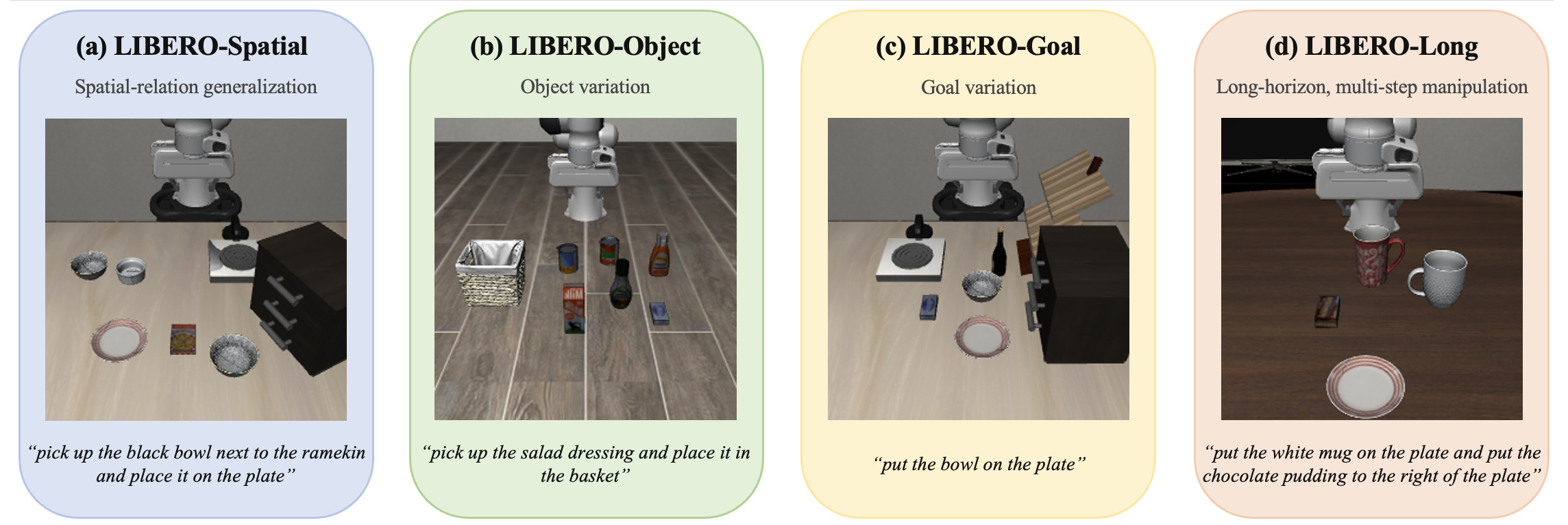}
\caption{\textbf{The four LIBERO evaluation suites}~\cite{libero}. Representative RGB observations and language instructions illustrate spatial relations (Spatial), object variation (Object), goal variation (Goal), and multi-step manipulation (Long). These panels introduce the task settings used in Table~\ref{tab:main}.}
\label{fig:libero-tasks}
\end{figure*}

\begin{figure*}[t]
\centering
\includegraphics[width=\textwidth]{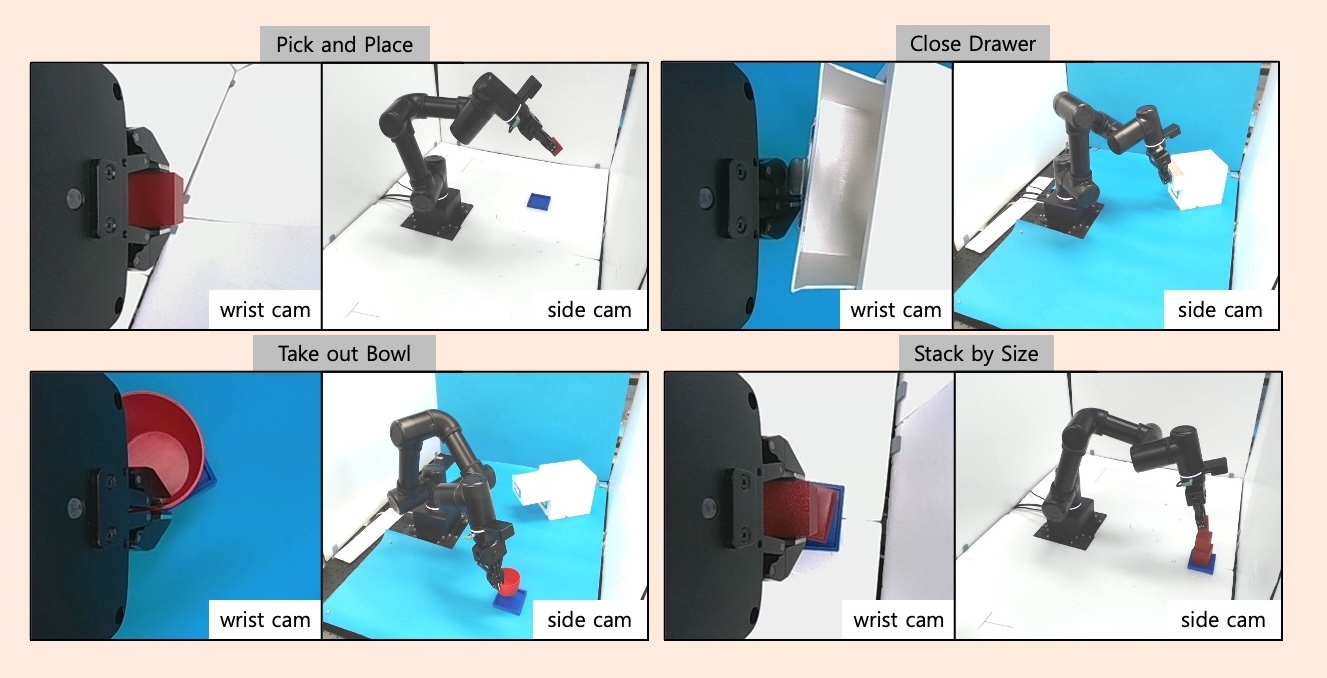}
\caption{\textbf{ROBOTIS OMY task settings.} Each task panel pairs the wrist camera (left) with the external camera (right). Top row: T1, red-cube pick-and-place; T2, drawer closing. Bottom row: T3, bowl retrieval and transfer; T4, size-ordered block stacking. The photos document camera views and manipulation settings. Table~\ref{tab:omy-real-success} reports task success rates for SmolVLA and \method{}.}
\label{fig:omy-tasks}
\end{figure*}

\subsection{Stage II: Geometry-Guided Action Learning}

\textbf{Geometric residual adaptation.}
We freeze the bridge and geometry decoder and learn an action-facing
interface. The feature map $g$ is pooled to the native patch grid and
projected by a learnable $1\times1$ convolution $R_\omega$:
\begin{equation}
\begin{aligned}
\Delta z &= \alpha R_\omega(\operatorname{Pool}(g)),\\
\widetilde z &= z+\Delta z,\qquad \alpha=\sigma(\eta).
\end{aligned}
\label{eq:residual}
\end{equation}
The projection matches the dimensions of $z$, and the scalar gate $\alpha$
controls residual magnitude. We initialize the weights and bias of
$R_\omega$ to zero and set the initial gate value to
$\alpha_0=\sigma(\eta_0)\approx0.002$. Consequently,
$\Delta z=0$ and $\widetilde z=z$ at initialization, so Stage-II
optimization starts exactly from the native visual representation. 
The adapted tokens enter the existing visual conditioning pathway, providing 
geometry-enriched features without introducing additional tokens.

\textbf{Action optimization.}
We train the native action expert, action-side projections, and residual
interface using the backbone's flow-matching objective~\cite{flow} with a
residual penalty:
\begin{equation}
\begin{aligned}
\mathcal{L}_{\mathrm{II}}
&=\mathcal{L}_{\mathrm{flow}}
+\lambda_{\mathrm{res}}\mathcal{L}_{\mathrm{res}},\\
\mathcal{L}_{\mathrm{res}}
&=\sqrt{\operatorname{mean}(\Delta z^2)+\epsilon}.
\end{aligned}
\label{eq:action}
\end{equation}
Here $\lambda_{\mathrm{res}}$ is a fixed residual regularization weight and
$\epsilon$ is a small numerical constant. The residual penalty discourages
excessive changes to the native visual representation. The visual encoder,
visual projector, VLM, bridge, and geometry decoder remain frozen, while
gradients through the fixed conditioning pathway reach the residual
interface. This stage requires no depth supervision. Both LIBERO and OMY
follow this Stage-II optimization scheme; they differ in the source of
Stage-I depth targets and the action demonstrations used for Stage II.

\section{Experiments}
\label{sec:experiments}

We evaluate manipulation success on LIBERO and a physical ROBOTIS OMY
robot, depth prediction from the shared encoder, and the effect of disabling
the residual at inference. All policy experiments use SmolVLA; transfer to
other VLA backbones remains untested. The method preserves the native
visual-token count but adds a geometry decoder and residual interface.

\subsection{LIBERO Data and Training}
\label{sec:exp-setup}

\textbf{Geometry supervision.}
For Stage-I depth adaptation, we use a LIBERO-derived RGB-D collection with
metric-depth supervision. We train Stage I for 20,000 optimization steps with a batch size of 4. 

Consistent with Sec.~\ref{sec:method}, the native visual encoder remains
frozen during Stage I, while the feature bridge and geometry decoder are
optimized using Eq.~\ref{eq:depth}.

\textbf{Action learning.}
Policy training uses the HuggingFaceVLA LIBERO collection containing 1,693
demonstrations and 273,465 action frames. SmolVLA and GeoBridge-VLA use the same action data and policy-training
conditions.
Both policies are trained for 50,000 optimization steps with a batch size of 32.
Evaluation protocols are described in Sec.~\ref{sec:eval-protocols}.

Before Stage-II action learning, the feature bridge and geometry decoder are
initialized from Stage I and subsequently frozen. During Stage II, only the
residual interface, action-side projections, and native action expert are
optimized using Eq.~\ref{eq:action}. No depth supervision is used during this
stage.

At inference time, both methods receive RGB observations, robot state, and
language instructions. GeoBridge-VLA derives geometric features from the same
frozen visual hierarchy used by the policy and injects them through the
learned residual interface. Neither method receives depth as a rollout
observation.

\subsection{Evaluation Protocols}
\label{sec:eval-protocols}

\textbf{LIBERO manipulation.}
Figure~\ref{fig:libero-tasks} shows representative scenes from Spatial,
Object, Goal, and Long. We evaluate each policy on 40 tasks across
the four suites, with 10 tasks per suite and 50 episodes per task,
yielding 500 episodes per suite and 2,000 episodes in total.
We report success rates for each suite and their unweighted mean.
Both methods use the same evaluation conditions. This comparison measures
the performance of the complete adaptation pipeline. The residual
intervention below tests the effect of its action interface at inference.

\textbf{Depth prediction.}
We report AbsRel, RMSE in meters, and $\delta_1$ for the SmolVLA
shared-encoder model and DA3-Mono-Large on the same LIBERO front-view 
validation set.

\textbf{Residual intervention.}
We evaluate the same trained GeoBridge-VLA checkpoint with the residual
enabled (normal) and disabled (zero). The zero condition sets $\Delta z=0$
at inference without changing any model weights or retraining the policy.
We report $\Delta$ as normal minus zero, so positive values indicate higher
success when the residual is enabled. This intervention measures
inference-time sensitivity, not the effect of training without geometry.

\subsection{Real-World ROBOTIS OMY Setup}
\label{sec:omy-setup}

Table~\ref{tab:omy-instructions} lists the four OMY tasks and
the language instruction used for each task.

\begin{table}[t]
\caption{\textbf{OMY tasks and language instructions.}}
\label{tab:omy-instructions}
\centering\footnotesize
\setlength{\tabcolsep}{4pt}
\renewcommand{\arraystretch}{1.15}
\begin{tabularx}{\columnwidth}{@{}p{0.28\columnwidth}X@{}}
\toprule
Task & Language instruction \\
\midrule
T1: Cube transfer & Pick up the red cube on the right side and place it on the blue plate on the left side. \\
T2: Drawer closing & Close the drawer. \\
T3: Bowl transfer & Pick up the red bowl in the drawer and place it on the blue plate on the right side. \\
T4: Block stacking & Stack the three red cubes on the blue plate, from largest at the bottom to smallest at the top. \\
\bottomrule
\end{tabularx}
\end{table}

Figure~\ref{fig:omy-tasks} shows the wrist and external camera
views for the four physical task settings.

\textbf{Data collection and training.}
We collect 50 demonstrations per task using a ROBOTIS OMY robot,
recording front- and wrist-view RGB observations, robot states, and actions.
The task-specific SmolVLA baselines use the same collected action demonstrations.
OMY training follows the two-stage procedure in Sec.~\ref{sec:method}. Stage I is trained for 20,000 optimization steps with a batch size of 4, while Stage II is trained for 50,000 optimization steps with a batch size of 32. The task-specific SmolVLA baselines use the same batch size and number of optimization steps as Stage II for policy training.
In Stage I, we use Depth Anything 3 (DA3)~\cite{da3} as a frozen teacher to generate pseudo-depth targets for the collected RGB observations.
These targets supervise the feature bridge and K4 geometry decoder using
Eq.~\ref{eq:depth}, while the native SmolVLA visual encoder remains frozen.

In Stage II, we initialize the bridge and geometry decoder from Stage I and
freeze them. We then train the residual interface, including the front- and
wrist-view routing/readout components, together with the action-side
projections and native action expert on our directly collected robot
demonstrations. Optimization uses flow-matching loss and residual
regularization in Eq.~\ref{eq:action}. The visual encoder, visual projector,
and VLM remain frozen. Stage II uses no depth targets.
At deployment, the DA3~\cite{da3} teacher is removed, and both methods receive only RGB, robot state, and language.

\textbf{Evaluation.}
We evaluate each method over 50 matched autonomous trials per task. Each initial
configuration is shared between SmolVLA and GeoBridge-VLA, with the robot
start state, object placement, target placement, and task-specific initial
state reset before each rollout. Both methods use the same instruction,
camera setup, execution limits, and task-success criterion. A trial succeeds
only when the complete task is achieved without human intervention; an
incomplete task, timeout, or intervention is counted as a failure.

\section{Results}
\label{sec:results}

We report manipulation success, depth prediction, and residual-conditioning
results using the settings described in Sec.~\ref{sec:experiments}.
Real-robot evaluation is presented separately from simulation results.

\subsection{LIBERO Manipulation Performance}
\label{sec:libero-main}

\begin{table}[t]
\caption{\textbf{LIBERO success rates (\%).} Both methods use the same evaluation conditions. Overall is the suite average; $\Delta$ is the change in percentage points.}
\label{tab:main}
\centering\footnotesize
\setlength{\tabcolsep}{2.5pt}
\renewcommand{\arraystretch}{1.08}
\begin{tabularx}{\linewidth}{@{}>{\raggedright\arraybackslash}p{0.25\linewidth}>{\hsize=0.85\hsize\linewidth=\hsize\centering\arraybackslash}X>{\hsize=1.30\hsize\linewidth=\hsize\centering\arraybackslash}X>{\hsize=0.85\hsize\linewidth=\hsize\centering\arraybackslash}X@{}}
\toprule
\textbf{LIBERO Suite} &
\textbf{SmolVLA}\newline\textbf{(SR)} &
\textbf{\method{}}\newline\textbf{(SR)} &
\textbf{Change}\newline\textbf{($\Delta$, pp)} \\
\midrule
Spatial & 68.8\% & \textbf{73.4\%} & \textbf{+4.6} \\
Object  & 65.6\% & \textbf{86.4\%} & \textbf{+20.8} \\
Goal    & 72.8\% & \textbf{76.6\%} & \textbf{+3.8} \\
Long    & 32.8\% & \textbf{47.2\%} & \textbf{+14.4} \\
\midrule
\textbf{Overall} & 60.0\% & \textbf{70.9\%} & \textbf{+10.9} \\
\bottomrule
\end{tabularx}
\end{table}

\begin{figure}[t]
\centering
\includegraphics[width=\columnwidth]{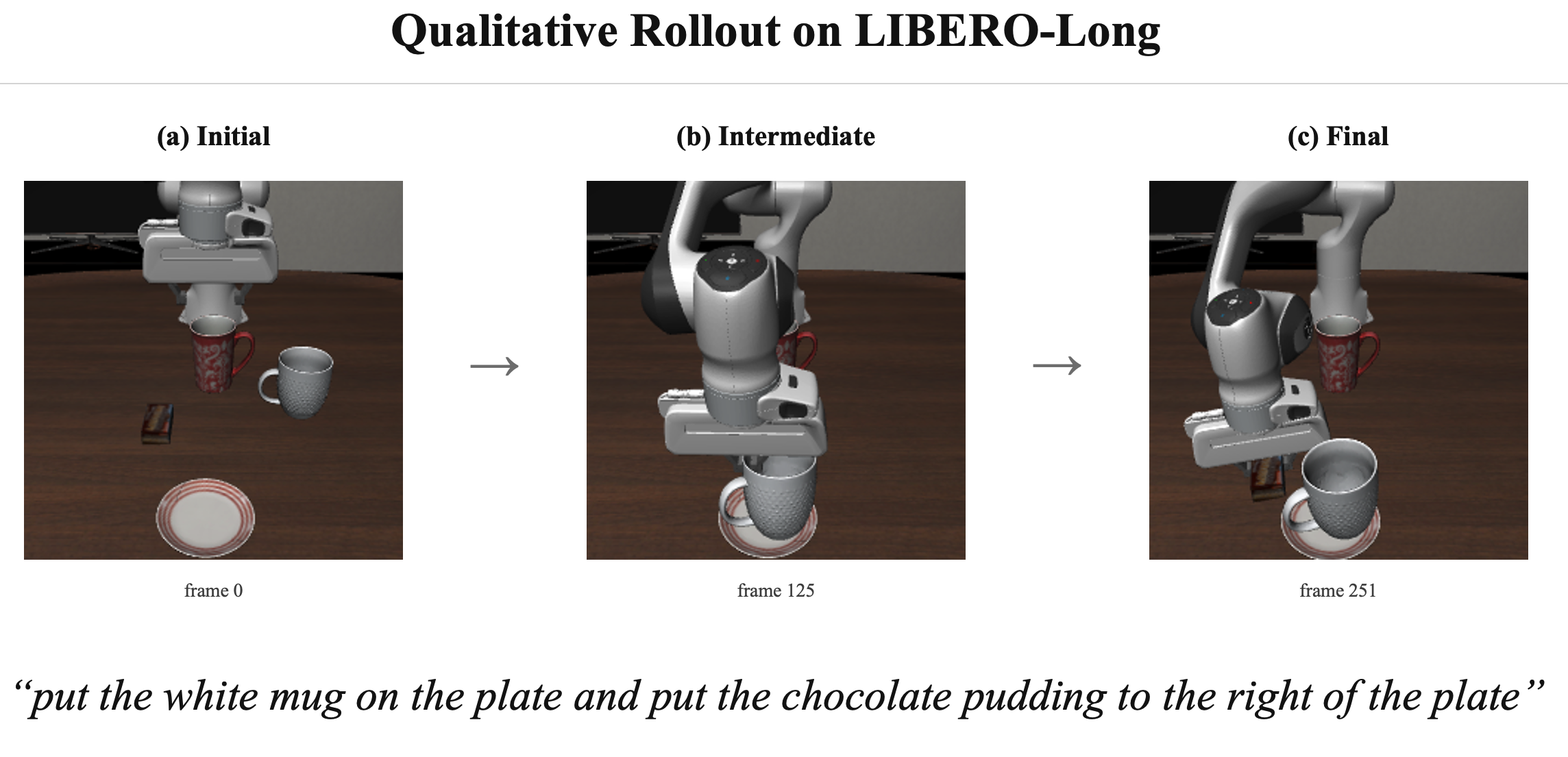}
\caption{\textbf{LIBERO-Long task sequence.} Frames 0, 125, and 251 illustrate progress through a multi-step mug-and-pudding manipulation task.}
\label{fig:libero-long}
\end{figure}

Table~\ref{tab:main} reports higher success for GeoBridge-VLA in all four
LIBERO suites, with an overall rate of 70.9\% compared with 60.0\% for
SmolVLA. The gains are largest on Object (+20.8 percentage points) and
Long (+14.4), followed by Spatial (+4.6) and Goal (+3.8). Both policies
use RGB observations, robot state, and language at inference, without
depth observations. Figure~\ref{fig:libero-long} shows representative frames
from a Long-suite rollout.

\subsection{Shared-Encoder Depth Prediction}
\label{sec:depth-analysis}

On the LIBERO front-view validation set, the shared-encoder model achieves
an AbsRel of 0.030, an RMSE of 0.1316\,m, and a $\delta_1$ of 0.975
(Table~\ref{tab:libero-depth-backbone}). The bridge and decoder recover
depth from the frozen policy encoder without an additional image encoder.

\begin{table}[t]
\caption{\textbf{Shared-encoder depth prediction.} Results on the LIBERO front-view validation set.}
\label{tab:libero-depth-backbone}
\centering
\footnotesize
\setlength{\tabcolsep}{2.5pt}
\begin{tabular*}{\columnwidth}{@{\extracolsep{\fill}}lrrr@{}}
\toprule
Model & AbsRel $\downarrow$ & RMSE (m) $\downarrow$ & $\delta_1$ $\uparrow$\\
\midrule
SmolVLA shared encoder & \textbf{0.030} & \textbf{0.1316} & 0.975\\
DA3-Mono-Large~\cite{da3} & 0.036 & 0.1549 & \textbf{0.988}\\
\bottomrule
\end{tabular*}
\end{table}

\begin{figure}[t]
\centering
\includegraphics[width=\columnwidth]{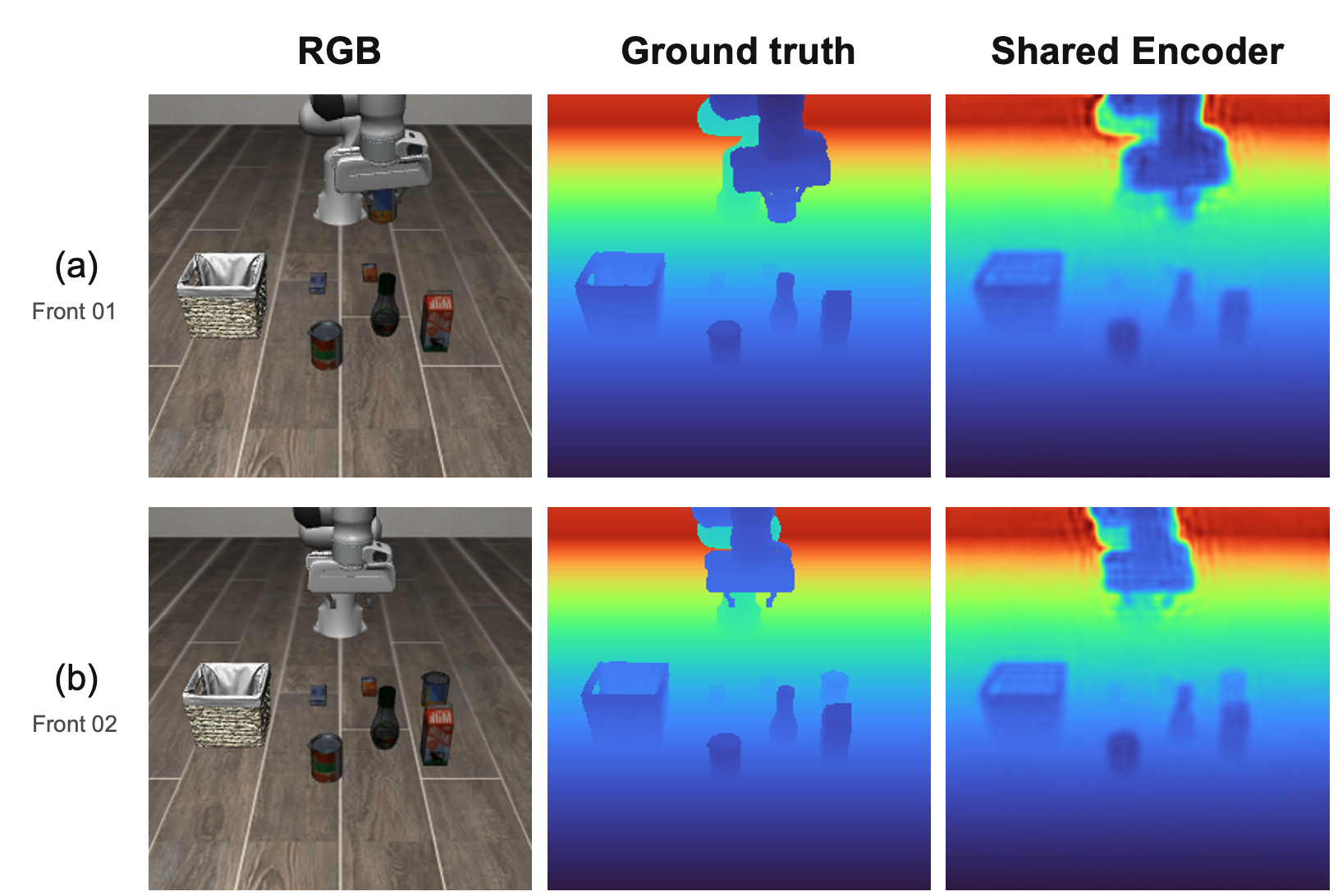}
\caption{\textbf{Shared-encoder depth predictions.} Qualitatively selected LIBERO front views show RGB (left), reference depth (middle), and predictions (right), illustrating robot, object, and tabletop geometry.}
\label{fig:shared-depth}
\end{figure}

Figure~\ref{fig:shared-depth} compares RGB observations, reference depth,
and shared-encoder depth predictions for two LIBERO front views,
complementing the quantitative results in Table~\ref{tab:libero-depth-backbone}.
The broad scene layout is recovered, while
fine object boundaries are visibly smoother than in the reference depth.

\begin{figure}[t]
\centering
\includegraphics[width=\columnwidth]{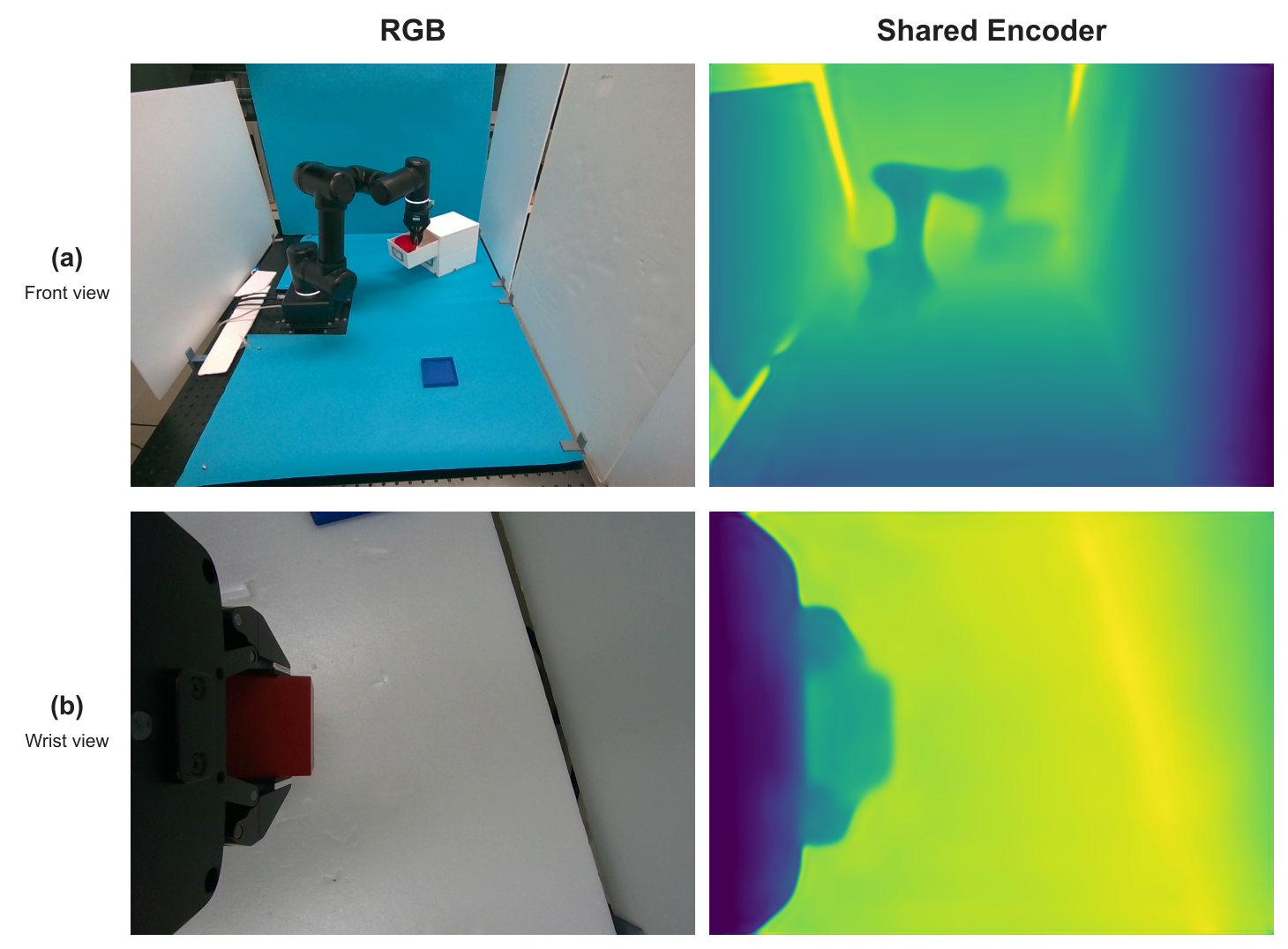}
\caption{\textbf{OMY shared-encoder depth predictions.} RGB observations (left) and predicted relative depth (right) for (a) the front view and (b) the wrist view. Examples are qualitatively selected from the training set.}
\label{fig:omy-depth}
\end{figure}

Figure~\ref{fig:omy-depth} pairs RGB observations with predicted relative
depth for the OMY front and wrist cameras, extending the qualitative
analysis to real-robot observations. The selected examples show the robot-arm
structure in the front view and the gripper-held object in the wrist view,
although fine boundaries remain smooth.

The \hyperref[app:decoder-results]{appendix} reports a separate comparison
of the standalone depth decoders.

\subsection{Residual Conditioning Analysis} 
\label{sec:residual-diagnostic} 

Table~\ref{tab:residual-ablation} compares GeoBridge-VLA with and without
the geometric residual at inference using the same trained checkpoint and the protocol in
Sec.~\ref{sec:eval-protocols}. This analysis examines the role of the
geometry-to-action interface beyond depth prediction.

\begin{table}[t]
\caption{\textbf{Same-checkpoint residual intervention (\%).} Normal and zero use the same trained checkpoint without retraining. Zero sets $\Delta z=0$ at inference; $\Delta$ is normal minus zero in percentage points; positive values indicate higher success with the residual enabled.}
\label{tab:residual-ablation}
\centering\footnotesize
\setlength{\tabcolsep}{3pt}
\begin{tabular*}{\columnwidth}{@{\extracolsep{\fill}}lrrr@{}}
\toprule
LIBERO Suite & Normal & Zero & $\Delta$ (pp)\\
\midrule
Spatial & 73.4\% & 70.2\% & +3.20\\
Object  & 86.4\% & 87.6\% & -1.20\\
Goal    & 76.6\% & 77.2\% & -0.60\\
Long    & 47.2\% & 44.4\% & +2.80\\
\midrule
Overall & 70.9\% & 69.85\% & +1.05\\
\bottomrule
\end{tabular*}
\end{table}

Enabling the residual yields 70.90\% overall success, compared with
69.85\% when it is disabled, a difference of 1.05 percentage points.
The residual improves Spatial and Long by 3.20 and 2.80 points,
respectively, but reduces Object and Goal by 1.20 and 0.60 points.
The effect therefore varies across suites. Because both conditions retain
the weights learned during Stage II, this experiment measures the effect
of enabling the residual at inference. It does not isolate the contribution
of geometry to policy learning.

\subsection{Real-World ROBOTIS OMY Success}
\label{sec:omy}

Table~\ref{tab:omy-real-success} reports results on cube transfer, drawer
closing, bowl transfer, and block stacking. Both policies use RGB, robot
state, and language and are evaluated from matched initial configurations
under the protocol in Sec.~\ref{sec:omy-setup}.

GeoBridge-VLA achieves 74.0\% overall success, compared with 54.0\% for
SmolVLA, an improvement of 20.0 percentage points. Gains range from 16.0
to 24.0 points across the four tasks. Block stacking shows the largest
increase, from 36.0\% to 60.0\%, but remains the most difficult task for
both policies. GeoBridge-VLA reaches 86.0\% on drawer closing and improves
cube and bowl transfer by 20.0 points each.

\begin{table}[t]
\caption{\textbf{OMY success rates (\%).} Each method is evaluated over 50 trials per task. Overall is the unweighted mean across the four tasks; $\Delta$ denotes GeoBridge-VLA minus SmolVLA in percentage points.}
\label{tab:omy-real-success}
\centering\footnotesize
\setlength{\tabcolsep}{3pt}
\begin{tabular*}{\columnwidth}{@{\extracolsep{\fill}}lrrr@{}}
\toprule
Task & SmolVLA & GeoBridge-VLA & $\Delta$ (pp)\\
\midrule
T1: cube transfer & 60.0\% & 80.0\% & +20.0\\
T2: drawer closing & 70.0\% & 86.0\% & +16.0\\
T3: bowl transfer & 50.0\% & 70.0\% & +20.0\\
T4: block stacking & 36.0\% & 60.0\% & +24.0\\
\midrule
Overall & 54.0\% & 74.0\% & +20.0\\
\bottomrule
\end{tabular*}
\end{table}

\section{Discussion} 
\label{sec:discussion}
Our results support three observations about the proposed geometry retrofit. 
First, the shared-encoder depth results show that depth-related geometric 
information can be decoded from intermediate representations already produced 
by the VLA visual encoder, without introducing a second image encoder. 
Together with the manipulation results, this supports the feasibility of 
using the native visual hierarchy as a substrate for geometry adaptation and 
subsequent action conditioning. 

Second, the improvement of the complete GeoBridge-VLA pipeline should be 
distinguished from the effect of the residual interface at inference. 
GeoBridge-VLA improves overall LIBERO success from 60.0\% to 70.9\%, 
whereas disabling the residual in the same trained checkpoint reduces overall 
success from 70.90\% to 69.85\%, with mixed effects across suites. Because 
the zero-residual condition retains the weights learned with the geometry 
pathway, this intervention measures the trained policy's inference-time 
sensitivity to the geometry-to-action interface rather than isolating the 
contribution of geometry during policy learning. A separately trained 
no-geometry control would be needed to more directly assess the training-time 
contribution of the geometry pathway. 

Finally, our evaluation is limited to a single VLA backbone, SmolVLA, and 
one physical robot platform, ROBOTIS OMY. The results characterize 
GeoBridge-VLA under these settings but do not establish that the same gains 
will transfer to other VLA architectures or robot embodiments. For the 
physical-robot setting, Stage-I adaptation relies on teacher-generated 
pseudo-depth, and sensitivity to the choice and quality of this supervision 
remains untested. Evaluating the retrofit strategy across additional 
backbones, robot morphologies, depth supervision sources, and manipulation 
domains is an important direction for future work.

\section{Conclusion}
\label{sec:conclusion}

We presented GeoBridge-VLA, a two-stage method that adapts frozen VLA
visual features with depth supervision and transfers the resulting geometric
representation to action prediction through a gated residual interface.
The deployed policy reuses its native visual encoder and requires no depth
observations.

GeoBridge-VLA improves overall LIBERO success from 60.0\% to 70.9\% and
OMY success from 54.0\% to 74.0\% relative to SmolVLA. These results
demonstrate the feasibility of retrofitting depth-supervised geometric
information into an existing VLA visual pathway without introducing a
second image encoder at deployment.

\FloatBarrier
\bibliographystyle{IEEEtran}
\bibliography{references}

\appendix[SPACE-CLIPv2 Geometry Decoder]
\label{app:decoder}

This appendix describes the decoder internals and standalone depth results.
The main method concerns connecting this decoder to native VLA features and
transferring its geometric representation through a residual interface.

\subsection{Fixed Local Token Aggregation}
Building on SPACE-CLIP's semantic--structural decoding framework~\cite{spaceclip},
the K4-based geometry decoder combines deep semantic context with shallower
structural features. K4 denotes aggregation from four fixed sampling
positions. At each aggregation stage, a projected bridge-feature map $T$ is
sampled at these positions relative to a spatial query $q$:
\begin{equation}
\begin{aligned}
r(q) &= \sum_{k=1}^{4}w_k(q)\,
\mathcal{S}_{\mathrm{bil}}(T,q+\delta_k),\\
\boldsymbol{w}(q) &= \operatorname{softmax}(W_wu(q)).
\end{aligned}
\label{eq:k4}
\end{equation}
Here $u(q)$ is the current decoder feature, $\delta_k$ is a fixed offset, and
$W_w$ produces four mixing weights. $\mathcal{S}_{\mathrm{bil}}(T,p)$ samples
a feature vector at $p$ by bilinearly interpolating the four surrounding grid
features. Sampling positions remain fixed, whereas mixing weights depend on
the decoder feature. The aggregated feature is transformed and added to
$u(q)$ through a gated residual connection.

\subsection{Standalone Depth Comparison}
\label{app:decoder-results}
SPACE-CLIP V1 and SPACE-CLIPv2 use the same frozen CLIP ViT-B/16 encoder,
the same training pairs, and the same task-disjoint image test
split. Table~\ref{tab:standalone-depth} compares the two configurations.
SPACE-CLIPv2 reduces AbsRel from 0.05239 to 0.04321 and RMSE from
0.06134\,m to 0.04993\,m, while increasing $\delta_1$ from 0.95497 to 0.96678.
These improvements support the fixed-local decoding configuration under
the matched backbone and data settings, rather than establishing a separate
action-policy gain. The SmolVLA diagnostic in
Table~\ref{tab:libero-depth-backbone} uses a different validation population;
its absolute metrics do not define a controlled cross-backbone ranking.

\par\smallskip
\noindent\begin{minipage}{\columnwidth}
\captionof{table}{\textbf{Standalone depth comparison.} Matched CLIP backbone and 9,984-image LIBERO test split.}
\label{tab:standalone-depth}
\centering\footnotesize
\setlength{\tabcolsep}{2.5pt}
\begin{tabular*}{\columnwidth}{@{\extracolsep{\fill}}lrrr@{}}
\toprule
Decoder & AbsRel $\downarrow$ & RMSE (m) $\downarrow$ & $\delta_1$ $\uparrow$\\
\midrule
SPACE-CLIP V1 & 0.05239 & 0.06134 & 0.95497\\
SPACE-CLIPv2 & \textbf{0.04321} & \textbf{0.04993} & \textbf{0.96678}\\
\bottomrule
\end{tabular*}
\end{minipage}
\par\smallskip
\FloatBarrier
\end{document}